\documentclass{article}

\usepackage[preprint, nonatbib]{neurips_2019}

\usepackage[utf8]{inputenc} 
\usepackage[T1]{fontenc}    
\usepackage{hyperref}       
\usepackage{url}            
\usepackage{booktabs}       
\usepackage{amsfonts}       
\usepackage{amsmath}       
\usepackage{nicefrac}       
\usepackage{microtype}      
\usepackage{xcolor}
\usepackage{mathabx}
\usepackage{amsthm}
\usepackage{soul}
\usepackage{dirtytalk}
\usepackage{graphicx} 
\usepackage{bmpsize}
\usepackage{url}
\usepackage{subcaption}
\usepackage{wrapfig}

\title{Anomaly scores for generative models}
\author{
    Václav Šmídl\\
    Department of Computer Science\\
    Czech Technical University in Prague\\
    \texttt{smidl.vaclav@fel.cvut.cz}\\
    \And
    Jan Bím\\
    Department of Computer Science\\
    Czech Technical University in Prague\\
    \texttt{bim.jan@fel.cvut.cz}\\
    \And
    Tomáš Pevný\thanks{Tomáš Pevný is also with Cisco Systems, Inc.}\\
    Department of Computer Science\\
    Czech Technical University in Prague\\
    \texttt{pevnytom@fel.cvut.cz}
}

\begin{document}

\maketitle

\begin{abstract}
  Reconstruction error is a prevalent score used to identify anomalous samples when data are modeled by generative models, such as (variational) auto-encoders or generative adversarial networks. This score relies on the assumption that normal samples are located on a manifold and all anomalous samples are located outside. Since the manifold can be learned only where the training data lie, there are no guarantees how the reconstruction error behaves elsewhere and the score, therefore, seems to be ill-defined. This work defines an anomaly score that is theoretically compatible with generative models, and very natural for  (variational) auto-encoders as they seem to be prevalent. The new score can be also used to select hyper-parameters and models. Finally, we explain why reconstruction error delivers good experimental results despite weak theoretical justification.
\end{abstract}

\section{Motivation}
\label{sec:motivation}
While supervised classification and generative models have made tremendous progress thanks to neural networks in past years, little has been done in applications of neural networks in detecting anomalous or novel samples.\footnote{By anomalous / outlying / novel samples this text understands samples that differ from the majority in such manner that they raise a suspicion that they have been generated by some other distribution.} It has been recently shown in~\cite{vskvara2018generative} that vanilla k-nearest neighbor with basic $L_2$ distance is frequently superior to approaches based on Variational Auto-Encoders~\cite{kingma_auto-encoding_2013,ramachandran2017swish} (VAE) and Generative Adversarial Networks~\cite{goodfellow2014generative,schlegl2017unsupervised} (GAN). We believe that there are two problems with anomaly detection: (i) it is difficult to tune hyper-parameters and architectures of neural networks without known examples of anomalies, and (ii) scores used to identify anomalous samples are not well theoretically founded or they rely on assumptions which are difficult to assert in practice. This text, therefore, \emph{does not propose another architecture for anomaly detection}, but it focuses on the second problem --- identifying a good score for anomaly detection with (variational) autoencoders. Although the score is derived for VAEs, it is compatible with GANs as well.

\begin{figure}[t]
\begin{subfigure}{0.32\textwidth}
\includegraphics[width=0.95\linewidth]{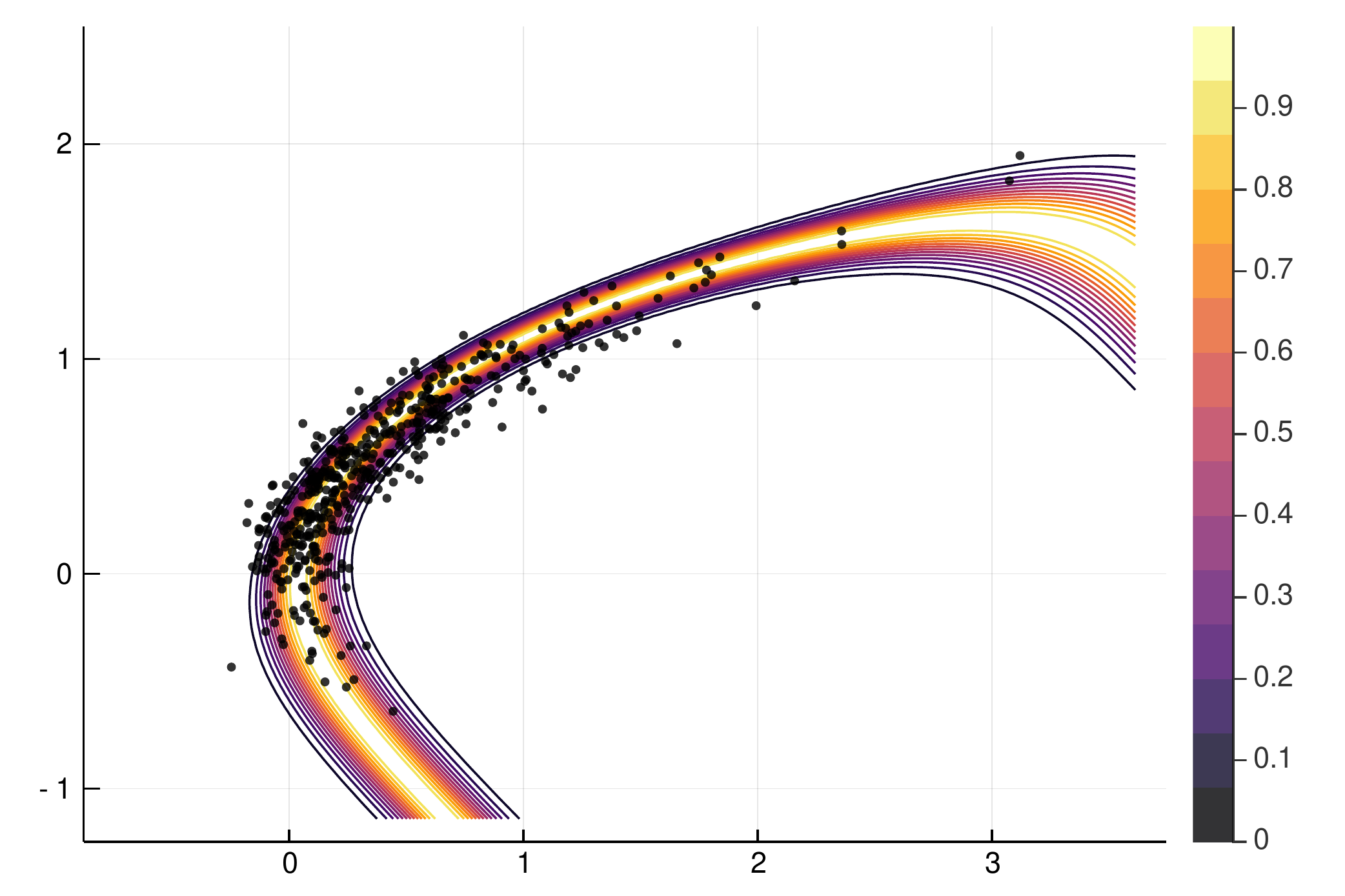}
\caption{Reconstruction error}
\label{fig:pxvita}
\end{subfigure}
\hfill
\begin{subfigure}{0.32\textwidth}
\includegraphics[width=0.95\linewidth]{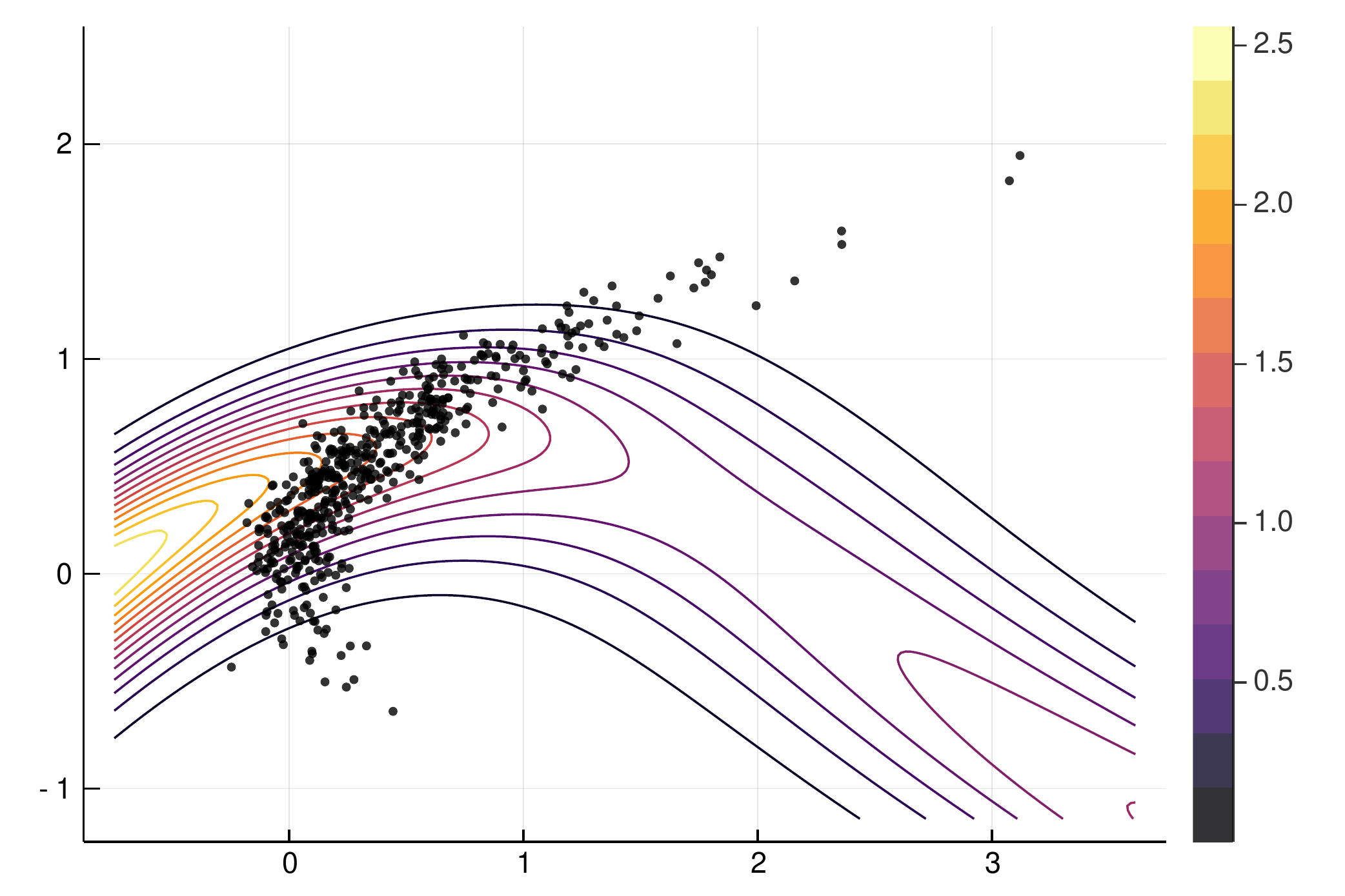}
\caption{Likelihood in latent space $p(z)$}
\label{fig:pz}
\end{subfigure}
\hfill
\begin{subfigure}{0.32\textwidth}
\includegraphics[width=0.95\linewidth]{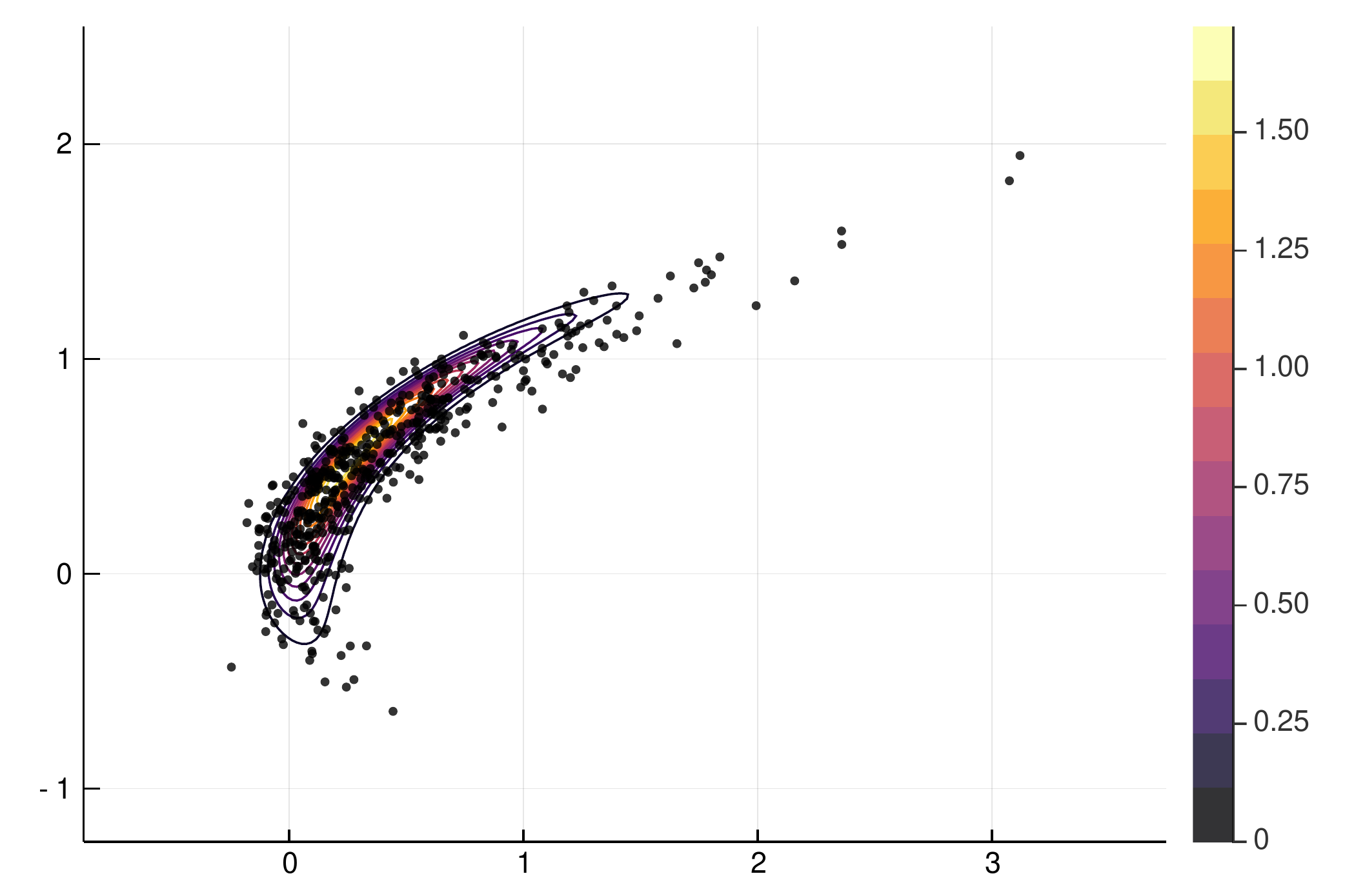}
\caption{The proposed score}
\label{fig:pxz}
\end{subfigure}
\caption{\label{fig:1}Anomaly scores of three different approaches for a toy problem, $x=[z^2,z]^\top+e$, where $p(z)= \mathcal{N}(0.5,0.15)$ and $p(e)=\mathcal{N}([0,0]^\top,0.01  \mathbf{I})$. (a) reconstruction error $p_{RE}(x)=\exp\left(-\frac{1}{2}||x-f(g(x))||_2^2/\sigma_{RE}\right), \sigma_{RE}=\mathrm{var}(x-f(g(x)))$, (b) $p(z)|_{z=g(x)}$, and (c) the proposed score (\ref{eq:p(x)_jacodeco}).}
\end{figure}

The most popular score for anomaly detection with neural networks is the reconstruction error of auto-encoders~\cite{an2015variational,zhou2017anomaly,xu2015learning} (although some works utilized it with restricted Boltzmann machines~\cite{fiore2013network}). The earliest use of this score, known to us, dates back to works of~\cite{bulitko2000using} and~\cite{hawkins2002outlier}. The theoretical justification of the reconstruction error as an anomaly score is that the data in the high-dimensional input space are located on some manifold $\mathcal{M}$. The autoencoder (replicator) network is trained to replicate the training data on the manifold, and if one further assumes that anomalous samples are located out of the manifold, they should have a high reconstruction error. Although the score seems reasonable, the problem is that it is very difficult to assert that reconstruction error is high in areas outside the manifold (this problem has been also documented in~\cite{nalisnick2018deep}). This problem is demonstrated in Figure~\ref{fig:pxvita}, where the reconstruction error of the autoencoder network defines the manifold well but also assigns high probability in areas where no data have been observed. We believe that previously reported positive experiences with reconstruction error is caused by sheer luck that anomalies avoided such areas.

An alternative to reconstruction error is to model distribution in the latent space (image of the encoding function), used for example in~\cite{zong2018deep}, and deem anomalies as samples with low probability in the latent space. The inherent problem of this score, demonstrated in Figure~\ref{fig:pz} is that due to the many-to-one relation of the encoder, the areas outside the manifold received probability that is too high.

A natural remedy to the problems identified above would be to somehow combine both scores together. In Section~\ref{sec:scores} it is argued that doing so with widely adopted generative models requires evaluation of a high-dimensional integral. This paper focuses on an anomaly score for generative models that (i) is theoretically justified, (ii) it combines advantages of reconstruction error and probability of a sample in the latent space, and (iii) it has reasonable computational complexity. The score (shown in Figure~\ref{fig:pxz}) utilizes a different re-parametrization of the generative model, which allows the calculation of the \emph{exact likelihood}. Moreover, it can be used with both variational autoencoders and generative adversarial networks. An experimental evaluation on problems from~\cite{vskvara2018generative} confirms its quality.

\subsection{Problem definition and notation.}
This work assumes that the data lie in some high-dimensional space $\mathcal{X} = \mathbb{R}^d.$ It is assumed that they are generated by a noisy projection $f_{\mathcal{M}}: \mathbb{R}^k \rightarrow \mathbb{R}^d$ as $\left\{f_{\mathcal{M}}(z) + e \vert z \sim p(z), e \sim p_e\right\},$ where $p(z)$ is some probability density on $\mathcal{Z} = \mathbb{R}^k$ and $p_e$ is a noise distribution. The goal is to estimate probability distribution $p(x)$ of the data given set of samples $\{x^{(i)}\}_{i=1}^{n}$ and assuming some $p(z)$ allowing simple calculation of the likelihood.

The proposed score relies on an ability to estimate $f_{\mathcal{M}},$ which is done by a parametric approximation by $f_{\theta}:\mathbb{R}^k \rightarrow \mathbb{R}^d$ with parameters $\theta$. Since $f_{\theta}$ is difficult to invert, the pseudoinverse is approximated by another parametric function $g_{\phi}:\mathbb{R}^d \rightarrow \mathbb{R}^k$ with parameters $\phi.$ Both functions $f_{\theta}$ and $g_{\phi}$ are typically implemented by some type of neural network. 
If it is clear from the text, the dependency on parameters $\theta, \phi$ is dropped.

\section{Variational autoencoders and Generative Adversarial Networks}
\label{sec:vae}
In this section, we briefly recapitulate generative models used in definitions of anomaly score reviewed or introduced in Section~\ref{sec:scores}.

\subsection{Variational Autoencoders}
Variational autoencoders~\cite{kingma_auto-encoding_2013} are probabilistic models representing distribution of observations $p(x)$ via a flexible transformation of the latent variable
$z.$ The assumed generative model is 
\begin{equation*}
p(x)=\int p_{\theta}(x|z)p_{\theta}(z)dz,
\label{eq:vae-gener}
\end{equation*}
where distribution $p_{\theta}(x|z)$ known as the \emph{decoder} is
$p_{\theta}(x|z)=\mathcal{N}(x;f_{\theta}(z),\sigma^2 \mathbf{I}),
\label{eq:vae-lik} $
with $p_{\theta}(z)$ being a known prior
on the latent variable and $f_{\theta}(z)$ being a neural network parametrizing
the mean of the Normal distribution. To
estimate parameters $\theta$ of $f_{\theta}(z)$ and possibly of the prior $p_{\theta}(z)$~\cite{tomczak2017vamp} from set of observations, $\{x^{(i)}\}_{i=1}^{n}$,
VAE introduces \emph{encoder}, which is a conditional probability distribution
$q_{\phi}(z|x)$ parametrized similarly to the decoder as $q_{\phi}(z|x) = \mathcal{N}(z;g_{\phi}(x),\mathbf{I})$. During training, parameters  $(\theta,\phi)$ are obtained by minimizing KL-divergence between $p_{\theta}(x|z)p_{\theta}(z)$ and $\frac{1}{n}\sum_{i=1}^n q_{\phi}(z|x^{(i)})$.


\subsection{Wasserstein Autoencoders}
\label{subsec:wae}
The original VAE minimizes the KL-divergence, which is effective if it can be computed analytically. However, it can be cumbersome in other cases, for example in the case of von Mises distributions in the latent space as used in the experiments' section. An alternative is to replace the KL-divergence with Wasserstein distance~\cite{tolstikhin2017wasserstein} \textbf{}(or any other distance of probability measures). The loss function minimized during training is then
\begin{equation}
 \min_{\theta,\phi}\frac{1}{n}\sum_{i=1}^{n}\log p_{\theta}(x^{(i)}|g_{\phi}(x^{(i)})) +\beta I(\{g_{\phi}(x^{(i)})\}_{i=1}^{n}||\{z_{j}\sim p_{\theta}(z)\}_{j=1}^{n}),\label{eq:vae-loss-wasser}
\end{equation}
where $\beta$ is a tuning parameter (corresponding to variance of the observation noise) and $I$ is a (Wasserstein) distance or a divergence.
The main advantage is that unlike the KL divergence, Wasserstein distance is estimated from two sets of points, and therefore it can better match the posterior to the prior~\cite{tolstikhin2017wasserstein}. The experiments below use Maximum Mean Discrepancy (MMD) \cite{borgwardt2006integrating} with IMQ kernel~\cite{tolstikhin2017wasserstein} as a distance for its ease of use due to its convenient calculation.

\subsection{Generative adversarial networks}
The last class of methods to fit the above generative models are Generative Adversarial Networks (GANs)~\cite{goodfellow2014generative}. Unlike previous models, in their basic variant they do not have an encoder network $g_{\phi}(x).$ Instead, they train a discriminator $t_{\omega}(x):\mathcal{X} \rightarrow \mathbb{R}$ implemented as a neural network with parameters $\omega$. Its goal is to learn the difference between true samples $X$ and samples generated by the generative model $p_{\theta}(x|z)$. During optimization, the goal is to find a saddle point of the following optimization problem
\begin{equation}
\min _{\theta}\max_{\omega} \frac{1}{n}\sum_{i=1}^n\left[\log t_{\omega}(x^{(i)}) + \log (1-t_{\omega}(p_{\theta}(x|z^{(i)})))\right], 
\end{equation}
where $\{x^{(i)}\}_{i=1}^{n}$ are the training data and $\{z^{(i)}\}_{i=1}^{n}$ are samples from $p(z)$. Originally GANs lacked the inference network (encoder) $q(z|x),$ but some variants~\cite{ulyanov2018takes,yi2017dualgan,zhu2017unpaired} incorporated it, by which they brought GANs closer to VAEs.

\section{Anomaly scores}
\label{sec:scores}
The established definition of an anomaly is \emph{an event occurring with a probability so low, that it raises suspicion of being generated by some other probability distribution}. This implies that a sample $x$ is an anomaly if the probability density function $p(x)$ is very low. Consequently, an anomaly score universal across a wide range of application domains\footnote{Some definitions of the anomaly are domain-specific and therefore they are not considered here.} should be proportional to the \emph{sample likelihood}.  Since anomaly detectors are evaluated on a range of probability density levels (e.g. via the AUC criteria), it is not necessary to compute the normalization constant. Therefore, we prefer to use term \emph{score} to denote unnormalized probability density.

\subsection{Evaluation of sample likelihood}
While the vast majority of literature on generative models is concerned with sampling from $p(x)$, there is not a generally accepted measure for evaluation of the probability $p(x)$, neither a corresponding anomaly score. 
For the generative models recapitulated in the previous section, to estimate probability of a sample $p(x)$ one should marginalize $p(x,z),$ which amounts to calculating an integral 
\begin{equation}
p(x)=\int p_{\theta}(x|z)p(z)dz=E_{z\sim q(z|x)}\left[p_{\theta}(x|z)\frac{p(z)}{q(z|x)}\right].\label{eq:px-correct}
\end{equation}
This is in a sharp contrast to the prior art on the use of autoencoders in anomaly detection (see Section~\ref{sec:related}). Autoencoders use the reconstruction error corresponding to the following unnormalized probability density
\begin{equation}
p(x)\approx\int p_{\theta}(x|z)\delta(z-g_{\phi}(x))dz\ \propto  \  \exp\left(-\|x - f(g(x))\|^2\right),
\label{eq:px-vita}
\end{equation}
where $\propto$ denotes equality up to a multiplicative constant. Note that 
in this score, the integration over $z$ is replaced by an evaluation of the likelihood at the "most probable" point given by the encoder $q_{\phi}(z|x).$  A different approach is used with the flow-based models, such as \cite{papamakarios2017masked}, where the the probability is evaluated by the change of variables formula
\begin{equation}
p(x)=p_{z}(f^{-1}(x))\left|\frac{\partial f^{-1}(x)}{\partial x}\right|.
\label{eq:px-J}
\end{equation}
However, there are some unexplained phenomena reported in~\cite{nalisnick2018deep} when this formula is used. Moreover, auto-regressive flows can be used only when the dimension of $\mathcal{Z}$ is equal to that of $\mathcal{X}$, i.e. $d=k$. While it is theoretically possible to map the redundant dimensions to the latent space, the training of such model is much harder. 

\subsection{The proposed alternative}
A common assumption in unsupervised models~\cite{alain2014regularized,vincent2008extracting,hinton2006reducing} is that the data observed in a high-dimensional input space are up to some noise located on some low-dimensional manifold $\mathcal{M}\subset\mathcal{X}$. It means that there is a function $f_{\mathcal{M}}:\mathcal{Z} \rightarrow \mathcal{X},$ such that $\mathcal{M} = \{f_{\mathcal{M}}(z) \vert z \in \mathcal{Z}\}.$ Furthermore it has been show in~\cite{dai2019diagnosing} that if the latent space $\mathcal{Z}$ has a lower dimension than the input space, the standard VAE learns just the manifold, and only when both dimensions are equal, it fits the chosen distribution $p(z).$ The assumption of existence of the manifold $\mathcal{M}$ and a data generating function $f_{\mathcal{M}}$ is adopted in the proposed score.

In practice, data do not lie exactly on the manifold, as they are subject to some noise. An observed point $x$ can be therefore modeled as a point $x'\in\mathcal{M}$ and some residue $e,$ 
\begin{equation}
e=x-x'=x-f(z'),
\label{eq:residue}
\end{equation}
which is typically assumed to be Gaussian distributed with variance $\sigma^2$. This work assumes that the noise is isotropic but \emph{only} in the space $\mathcal{T}f(z')$ tangent to the manifold $\mathcal{M}$ at the point $f(z')$ and it is zero on the manifold. This assumption is very different from the prior art, where the noise is assumed to be isotropic in the full space $\mathcal{X}$ including the manifold. However, the assumption allows to factorize $p(x)$ as $p(x)=p(x')p(e),$ which further allows to express $p(x')$ through the transformation of variables (as in~\eqref{eq:px-J}) and after substitution it leads to
\begin{align}
p(x)=p(x')p(e)\ \propto\ p_{z}(f^{-1}(x'))\left|\frac{\partial f^{-1}(x')}{\partial x}\right|\mathcal{N}(x-x',\sigma^2  \mathbf{I}).
\label{eq:p(x)_jacodeco}
\end{align}
Here, we have used $p(e)=\mathcal{N}(0,\sigma^2 \mathbf{I})$ as if $e$ is defined on the full space $\mathcal{X}$. In this case, the redundant dimensions change only the multiplicative constant, since the noise is isotropic even in the rotation to coordinate of $e$ and the manifold.  If normalized probability is required, $p(e)$ has to be evaluated only in the relevant dimensions. This would be possible by optimization of the optimal projection point $z'$ in (\ref{eq:residue}), as it is discussed below. We leave this line of research for further study.

Notice that the proposed score uses reconstruction error~\eqref{eq:px-vita} popular in the prior art together with the exact likelihood used in~\eqref{eq:px-J}. It should, therefore, benefit from both scores and prevent pathological failures, as has been demonstrated on the motivational example in Figure~\ref{fig:1}. Moreover, unlike models based on auto-regressive flows, the proposed score is not restricted to cases when the input and the latent spaces have the same dimension.

If the auto-encoder was \emph{properly trained}, i.e. if $g(f(z) + e) = z,$ where $e \sim \mathcal{N}(0,\mathbf{\Sigma}_{f(z)})),$ then the above equation is equivalent to
\begin{equation}
p(x)=p(x')p(e) \approx p_{z}(g(x'))\left|\frac{\partial g(x')}{\partial x'}\right|\mathcal{N}(x-x',\sigma^2\mathbf{I}).
\label{eq:p(x)_jacoenco}
\end{equation}
$\mathcal{N}(0,\mathbf{\Sigma}_{f(z)}))$ above denotes a normal distribution with a non-zero variance only in the tangent space to $f(z).$
The proposed likelihood can be calculated either using encoder $g$ from~\eqref{eq:p(x)_jacoenco} or using decoder $f$ from~\eqref{eq:p(x)_jacodeco}. If it holds that $f$ and $g$ are inverses on the manifold, they can be used equivalently. Yet, according to experimental results the formulation~\eqref{eq:p(x)_jacodeco} delivers better results, which is attributed to the empirical findings that $f$ and $g$ are not proper inverses on the manifold.

\subsection{Analysis of the noise model}
Let's now briefly discuss the assumption of a noise $e$ in (\ref{eq:residue}) to have non-zero variance only in the space tangent to the manifold. It might seem to be very restrictive at the first sight, as it might not model the reality accurately. Below it is argued that since in practice one almost always fails to identify the true $f,$ the assumption does not decrease the expressiveness of the model, but it changes the structure of the noise. Models in Section~\ref{sec:vae} (and most prior art) assume a generative model $x = f_{\mathcal{M}}(z) + e, $ where $e \sim \mathcal{N}(0,\sigma^2\mathbf{I}).$ The noise can be decomposed into two parts $e_{\mathcal{M}}$ and $e_{\mathcal{M}^{\perp}}$ where the second part $e_\mathcal{M}^{\perp} \in \mathcal{T}f_{\mathcal{M}}(z)$ is tangent to a manifold $\mathcal{M}$ at $f_{\mathcal{M}}(z)$ and for the first part it holds that $f_{\mathcal{M}}(z) + e_{\mathcal{M}} \in \mathcal{M}.$ Due to the bijection of $f_{\mathcal{M}}$ there exists $z'$ such that $f_{\mathcal{M}}(z') = f_{\mathcal{M}}(z) + e_{\mathcal{M}}.$ Consequently, $x$ can be expressed as $x = f_{\mathcal{M}}(z') + e_{\mathcal{M}^{\perp}},$ which conforms with the model (\ref{eq:residue}). It effectively means that the part of the noise that is aligned with the manifold $\mathcal{M}$ is \emph{absorbed} by the distribution on the latent space $p(z)$ and by the learned decoder $f.$ This is due to the Theorem~1 in~\cite{dai2019diagnosing}, according to which there exists a continuous function $f$ such that the prior distributions $p(z)$ of both noise models can be the same almost everywhere. We believe (without a formal proof) that the structure of the noise of both models is likely to be different, but the effect in practice will be negligible. 

Finally, note that the formulation of the likelihood~\eqref{eq:p(x)_jacodeco} can be used with generative models identified by GANs. The point $z$ can be found by solving $\arg \min_{z} \|f(z) - x\|^2,$ however in practice one would probably use variants with encoders satisfying cyclic properties~\cite{zhu2017unpaired,ulyanov2018takes}, as they might be faster to solve the optimization problem and also more stable. 

\subsection{Determinant of the Jacobian}
The calculation of the Jacobian in the evaluation of $p(x')$ in~\eqref{eq:p(x)_jacodeco}, where $x' = f(z')$ seems to be an ill-posed problem, because the Jacobian $\frac{\partial f(z)}{\partial z}$ is a rectangular matrix due to $f:\mathbb{R}^k \rightarrow \mathbb{R}^d$ with $d > k.$ But recall that $x'$ always lays on the manifold, i.e. $f(z') = x' \in \mathcal{M},$ and therefore the determinant should be calculated only with respect to the coordinate system on the manifold, which is of dimension $k$ and therefore properly defined.

Let's align the coordinate system on $\mathcal{X}$ such that last $d - k$ coordinates spans the tangent space $\mathcal{T}f(z).$ Then due to the definition of the tangent space it holds that last $d - k$ columns of the Jacobian are all zeros, i.e. $(\forall j>k)\left(\left. \frac{\partial f(z)_j}{z_i} \right|_{z} = 0 \right).$ Contrary, first $k$ components of the Jacobian define a local approximation of $\mathcal{M}$ around the point $f(z).$ If $f$ is a bijection, which is assumed here, they have a non-zero determinant.

In practice, the determinant can be easily calculated using Singular Value Decomposition of $J(f(z)).$ Specifically, $\mathrm{svd}(J(f(z)) =  \mathbf {U\Sigma V^{*}},$ where $\mathbf{U}$ and $\mathbf{V}$ are unitary matrices and $\Sigma$ is a diagonal matrix with $k$ non-zero singular values on the diagonal. The columns of $V$ corresponding to zero singular values form a base of the tangent space $\mathcal{T}f(z)$), and those corresponding to non-zero singular values form a base of the local approximation of the manifold. Since determinant is equal to the product of eigenvalues and singular values are their square roots, the product of squares of non-zero singular values is equal to the determinant of $J(f(z))$ with $f(z)$ determining the manifold $\mathcal{M}$.

\subsection{Identifying the model}
During training, a pair of functions (encoder $g_{\phi}$ and decoder $f_{\theta}$) or their parameters need to be found such that (i) $f_{\theta}$ approximates the true manifold $f_{\mathcal{M}},$ (ii) $g_{\phi}$ minimizes the reconstruction error, i.e. $g(x) \approx \arg \min_{z \in \mathcal{Z}} \|x - f(z)\|^2,$ and (iii) $g(x)$ where $x \sim p(x)$ is similar to $z \sim p(z).$ In experiments presented below, this pair is found by minimizing loss of Wasserstein auto-encoders (\ref{eq:vae-loss-wasser}). This criterion has been preferred as according to~\cite{tolstikhin2017wasserstein} it optimizes $p(z)$ more directly than KL-Divergence. Moreover, unlike KL-Divergence, it allows one to treat $g_{\phi}(x)$ as a Dirac-impulse, which is rather problematic to achieve with KL-Divergence. Note that MMD can be replaced by any other measure of choice or by a trained detector as it is done in adversarial autoencoders~\cite{makhzani2015adversarial}.

 The trade-off between enforcing low reconstruction error and a good match of $p(z)$  is controlled by $\beta,$ which is equal to the variance of noise and it is treated as a hyper-parameter in the experiments below.

\section{Related work}
\label{sec:related}
Below, we briefly recapitulate the prior art on the use of neural networks to detect anomalies. A reader interested in other models is referred to an excellent survey~\cite{chandola2009anomaly} and experimental comparison in~\cite{pevny2016loda,emmott2013systematic}. 

The most popular approach to anomaly detection with neural networks relays on (variational) auto-encoders~\cite{an2015variational,zhou2017anomaly,xu2015learning} (AE) although some works utilized restricted Boltzmann machines~\cite{fiore2013network} (RBM) or generative adversarial networks~\cite{schlegl2017unsupervised} (GAN). 

The prior art employing auto-encoders and RBMs differs mainly in how they train the encoder-decoder pair, as regular auto-encoders are used in~\cite{bulitko2000using}, variational auto-encoders in~\cite{an2015variational}, contractive autoencoders in~\cite{monaco2016robust}, denoising auto-encoders in~\cite{xu2015learning,marchi2015novel}, and energy based models~\cite{zhai2016deep}. Relatively few work has been dedicated to the problem of the proper anomaly score, as most prior art assumes that the normal samples lie on some manifold and anomalies are located outside of it. Under this assumption, the reconstruction error makes sense, as the encoder $f$ should project anomaly to the nearest point on the manifold, and when reconstructed by the decoder the data would thus suffer a high reconstruction error. 

Ref.~\cite{zong2018deep} uses autoencoders to compress dimension of the input data, and then it fits a Gaussian mixture model on the latent representation $z = g(x)$ augmented by a "cycled" latent representation $z' = g(f(g(x)).$ Although the theoretical justification of the method is missing, it seems to deliver a good performance. The work is interesting as the anomaly score is defined in the latent space $z,$ where the mixture can be viewed as fitting a distribution $g(x), x \sim p(x).$

Since application of GANs to anomaly detection is not as straightforward as that of auto-encoders and their variants, there is a diversity of anomaly scores derived from the model. One of the first works~\cite{schlegl2017unsupervised} uses a combination of reconstruction error and the score of the detector in GANs. The rationale behind it is that GAN's detector should be able to recognize normal and anomalous data. But taking the quote\footnote{\say{For GANs, one thing to keep in mind is that the discriminator is not a generalized detector of weird things.  It is trying to tell whether a sample came from the real data or *one specific non-data distribution: the generator*. Because of that, it seems like the discriminator would only be useful for anomaly detection if you think you can make your generator resemble the anomalies you expect to need to detect.}} by Ian Goodfellow~\cite{goodfellow2017dogan} into the account, the use of GANs in outlier detection might not be as simple as plainly using the detector.

The likelihood of a sample $x$ to be sampled from $p(z)$ is used in~\cite{deecke2018anomaly}, where the algorithm tries to find a point $z$ with the smallest reconstruction error $\|x - z\|^2$ and uses the density $p(z)$ as an anomaly score. Since the generator is not a convex function, the search for $z$ is repeated several times and corresponding densities are averaged.

With respect to the prior art, this work is not concerned about how the model is trained, but how to properly define the anomaly score. The proposed score aims to be theoretically correct. It combines likelihood in the latent space $p(z)$ with the reconstruction error viewed as $p(x|x')$ with $x' = f(z)$ and the determinant of the Jacobian to accommodate the change of variables.

Lastly, as has been mentioned in Section~\ref{sec:scores}, flow-based models directly evaluate likelihood of $p_{\theta}(x).$ To our best knowledge, they have been utilized in few-shot learning~\cite{iwata2019supervised}, but not directly in anomaly detection. Their main drawback at the moment is that the dimension of the latent space has to be equal to that of the input space. A model proposed in this paper, where the noise is \emph{independent} on the latent random variable, offers an elegant solution and should be investigated in the future.

\section{Experiments}
\label{sec:experiments}
The experimental comparison of the proposed score to the reconstruction error has been performed on ten problems (breast-cancer-wisconsin, cardiotocography, haberman, magic-telescope, pendigits, pima-indians, wall-following-robot, waveform-1, waveform-2, yeast) adapted according to~\cite{emmott2013systematic,pevny2016loda} to anomaly detection. These problems have been also used in the study~\cite{vskvara2018generative} comparing sophisticated methods based on neural networks to k-nearest neighbor~\cite{harmeling2006outliers} and isolation forests methods~\cite{liu2008isolation}. The below study uses only \emph{easy} anomalies, as more difficult anomalous samples are located in areas of high densities of normal data, which raises doubts if they should be considered anomalous~\cite{emmott2013systematic}.
For each dataset, five distinct train/test splits were created. The data were split to 80\% of the dataset being used for training and the remaining 20\% for evaluation. The training set is contaminated with up to 10\% anomalies if they were present in the dataset. Obviously, these anomalies are not labeled and therefore they cannot be used during the training of models.

VAMP variation of variational auto-encoder~\cite{tomczak2017vamp} was preferred over the vanilla VAE~\cite{kingma_auto-encoding_2013}, because it offers more flexibility in the prior distribution $p(z),$ which is implemented as a mixture with tunable parameters. Similarly, the components in the mixture were von Mises-Fisher distributions~\cite{davidson_hyperspherical_2018} instead of the usual Normal distribution, as it has been experimentally shown in~\cite{Aytekin2018L2_VAE_anomaly} that mapping the latent space on a sphere yields higher performance.

For each dataset and split of the data, a large number of models (280 to be exact) were trained differing by hidden layer dimensions $\in \{32, 64\}$, latent layer dimension (if smaller or equal than the data dimension) $\in \{2, 4, 9\}$, number of components in the prior mixture $\in \{1, 4, 16\}.$ Both encoder and decoder contained three fully connected hidden layers of neurons with the "swish" activation function \cite{ramachandran2017swish}. In order to represent the data space well, the decoder contained an extra output layer of neurons with linear activation function. All models were optimized using ADAM optimizer~\cite{kingma2014adam} with default setting and with batch size 100 for 10000 steps. Finally, all experiments are implemented in Julia programming language~\cite{Julia} with Flux.jl~\cite{innes:2018}. The code for the experiments is available at \url{https://github.com/anomaly-scores/Anomaly-scores}. The closeness of distributions $g(x)$ where $x \sim p(x)$ and $z \sim p(z)$ is measured using Maximum Mean Discrepancy (MMD) with IMQ kernel~\cite{tolstikhin2017wasserstein}, where its width $c$ is treated as a hyper-parameter $c \in \{0.001, 0.01, 0.1, 1\}$. Finally, the trade-off between enforcing reconstruction error and closeness of the distributions represented by $\beta$ (see Equation~\eqref{eq:vae-loss-wasser}) is also treated as a hyper-parameter $\beta \in \{0.01, 0.1, 1, 10\}.$

\begin{figure}
    \centering
    \includegraphics[width=0.95\linewidth]{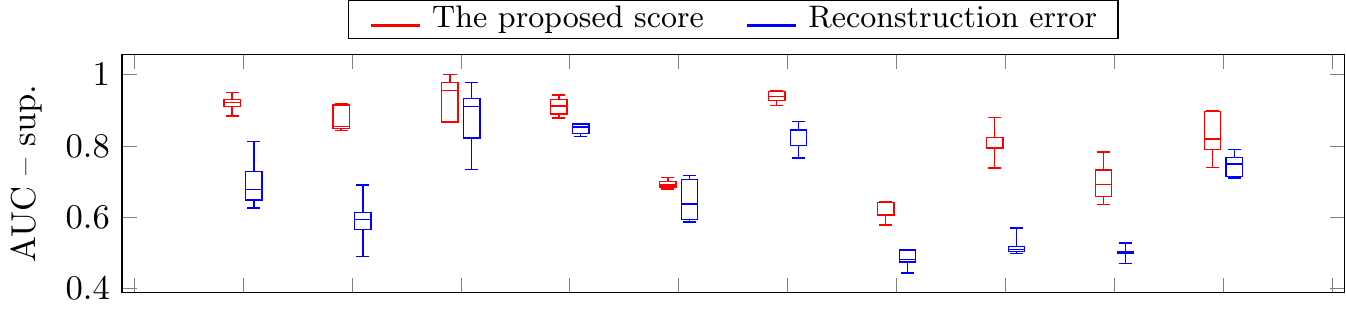}
    \includegraphics[width=0.95\linewidth]{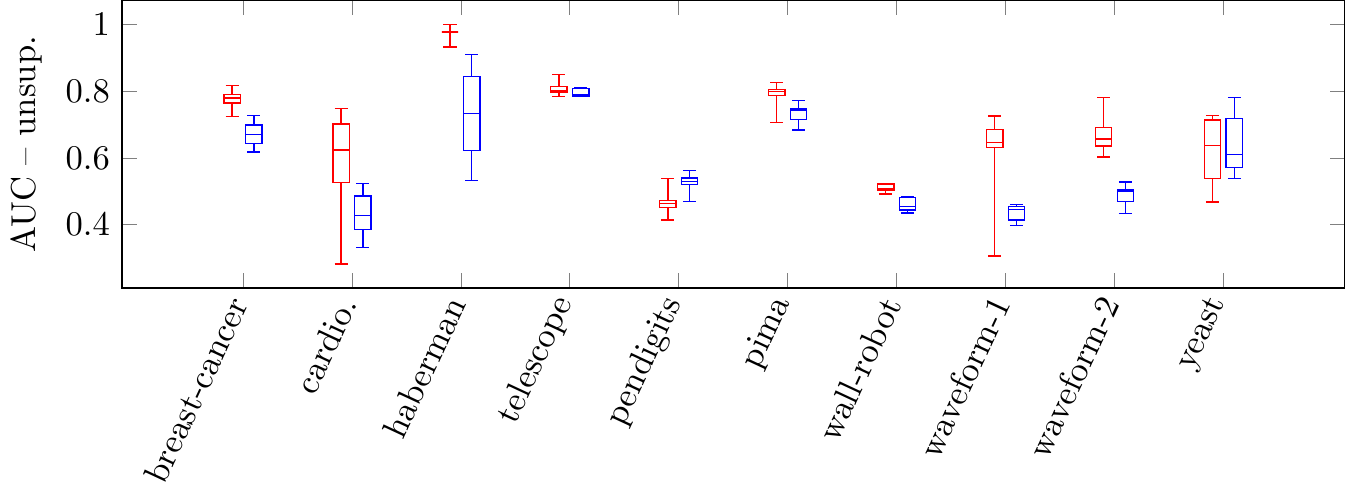}
    \caption{AUCs of the proposed score and reconstruction error of autoencoders with hyper-parameters selected according to supervised (top) and unsupervised (bottom) approach.}
    \label{fig:results}
\end{figure}\textbf{}
The quality of detection is measured using the area under the ROC curve (AUC), which is considered as a standard in the field of anomaly detection. Top figure~\ref{fig:results} shows AUCs for models utilizing reconstruction error and the proposed score in terms of box-plot. Since up to 280 models differing by hyper-parameters were trained for each problem and data-split, the best combination of hyper-parameters was selected according to AUC on the training set, which simulates the scenario where some examples of anomalies are available for model selection.\footnote{This scenario might not be as unrealistic as it sounds, as one typically has examples of a few anomalies.} The results show that the proposed score is better than the reconstruction error on all problems.

In the most strict scenario for anomaly detection, no labeled anomalies are available during training and even during model selection. The criterion to select the model is therefore important but rarely addressed problem. Bottom figure~\ref{fig:results} shows AUCs, where hyper-parameters were selected according to lowest reconstruction error on the training set for the score based on reconstruction error (lower is better), and according to the highest likelihood of the training data for the proposed score. Since the reconstruction error can be interpreted as a likelihood as well (see Equation~\eqref{eq:px-vita}), the model-selection criteria should be comparable. However, the selection is not accurate due to inaccuracy in the evaluation of the normalization constant mentioned in Section \ref{sec:scores}. The experimental results show that in this difficult setting the proposed score is better in nine out of ten problems. We believe that even better results can be obtained if the normalization constant is approximated more closely using the optimization of $z'$ as proposed in Section \ref{sec:scores}.

\subsection{Is there a manifold in the data?}
\begin{wrapfigure}{r}{0.5\textwidth}
    \centering
    \includegraphics[width=\linewidth]{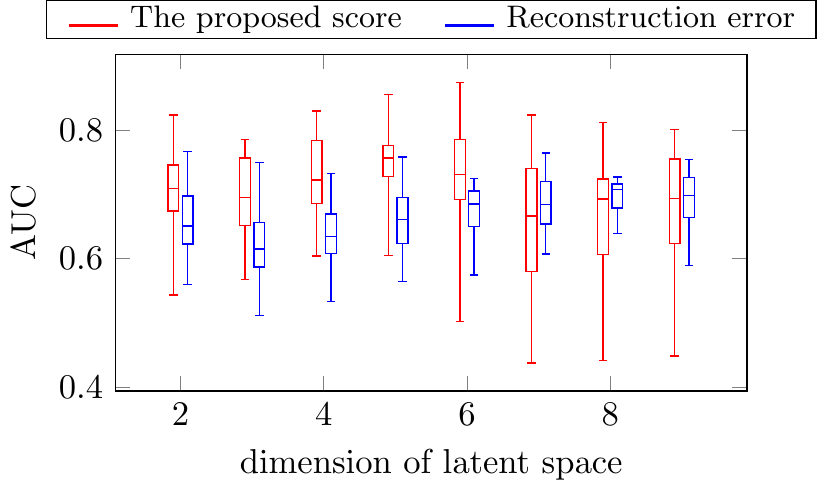}
    \caption{AUC of anomaly detection for a range of latent dimensions for the proposed criteria (red) and reconstruction error score (blue).}
    \label{fig:dimz}
\end{wrapfigure}Since the data lie in the full space, it should be possible to find a mapping to the same dimensional latent space, in a similar manner as the flow based methods. The decomposition into a manifold and noise part may be seen as a simplification of the full model. We test if this modeling assumption is valid on an exhaustive search over all possible dimensions of the latent space on the "breast-cancer-wisconsin" dataset which has eight dimensions. The results of anomaly detection for all possible latent dimensions are displayed in Figure \ref{fig:dimz} with variability with respect to the splits of test/train data and hyper-parameters.

Note that while the conventional score based on reconstruction error is almost insensitive to the latent dimension, the performance of the anomaly detection based on the proposed score has a flat peak at 4--6 dimensions with decreasing performance for lower as well as higher dimensions. This suggests, that the modeling assumption of the low-dimensional latent space is beneficial. This may be relevant to the discussion on the performance of the full-dimensional latent space \cite{nalisnick2018deep}.

\section{Conclusion}
This paper presented a new score to identify anomalies using deep generative models. The score utilizes a decomposition of the data space into manifold obtained by projection of the latent space and the residue. The assumption that the residue lies in the tangent space to the manifold allows combining the conventional scores based on reconstruction error with probability in the latent space and transformation of variables into a computationally efficient approach. The proposed score can be applied to variational autoencoders and GAN-based models (an application to flow-based models is trivial, as they lack the reconstruction term).

The experimental results demonstrated that the new score decisively outperforms conventional scores on anomaly detection, where hyper-parameters are selected using few labeled samples, and almost always outperforms them in unsupervised anomaly detection, where hyper-parameters are selected using likelihood. Since proper normalization is not required in anomaly detection, the paper utilized certain computational simplifications. In the paper, we have outlined, how to extend the approach for a better approximation of the exact probability density. We plan to pursue this direction further, as it might be important for selection of hyper-parameters. Other interesting direction would be an evaluation with other generative models, namely GANs and adversarial autoencoders.

\bibliographystyle{plain}
\bibliography{bibliography}

\end{document}